# Review on Causality Detection Based on Empirical Dynamic Modeling

Cao Zhihao , Qu Hongchun

College of Information Science and Engineering, Zaozhuang University

## Abstract

In contemporary scientific research, understanding the distinction between correlation and causation is crucial. While correlation is a widely used analytical standard, it does not inherently imply causation. This paper addresses the potential for misinterpretation in relying solely on correlation, especially in the context of nonlinear dynamics. Despite the rapid development of various correlation research methodologies, including machine learning, the exploration into mining causal correlations between variables remains ongoing. Empirical Dynamic Modeling (EDM) emerges as a data-driven framework for modeling dynamic systems, distinguishing itself by eschewing traditional formulaic methods in data analysis. Instead, it reconstructs dynamic system behavior directly from time series data. The fundamental premise of EDM is that dynamic systems can be conceptualized as processes where a set of states, governed by specific rules, evolve over time in a high-dimensional space. By reconstructing these evolving states, dynamic systems can be effectively modeled. Using EDM, this paper explores the detection of causal relationships between variables within dynamic systems through their time series data. It posits that if variable X causes variable Y (X $\Rightarrow$ Y), then the information about X is inherent in Y and can be extracted from Y's data. This study begins by examining the dialectical relationship between correlation and causation, emphasizing that correlation does not equate to causation, and the absence of correlation does not necessarily indicate a lack of causation. It then delves into the core principles of causality detection based on EDM, tracing the historical evolution of key concepts such as Takens' Embedding Theorem, the Simplex Projection algorithm, and the Convergent Cross Mapping algorithm. Further, the paper discusses advancements in EDM theory and its practical applications in causal detection. Finally, it offers insights into the prospective future trends in causal detection research based on EDM.

# 1. Introduction

Correlation is a crucial analytical criterion in contemporary scientific research. Researchers across various fields, including ecology and epidemiology, often focus on uncovering Correlation relationships between variables [1]. However, the rise of awareness regarding non-linear dynamics, especially in real-world scenarios and ecosystems, highlights a limitation: sole reliance on correlation for inference can lead to misleading conclusions. Phenomena such as "spurious correlations" and "illusory correlations" demonstrate that relationships in dynamic systems can unpredictably change. A positive correlation between two variables might become negative over time, indicating the instability of conclusions based only on correlation. This situation calls for a more reliable standard in scientific research [2].

As Berkeley articulated in his work *A treatise concerning the principles of human knowledge*, the simultaneous occurrence of events does not necessarily imply a causal relationship, implying that correlation does not equate to causation [3,4]. The famous Song dynasty poet Su Xun's line "月晕而风，础润而雨" (When there's a lunar halo, there will be wind; when the cornerstone is moist, there will be rain), illustrates this point. While lunar halos and rainy weather may appear highly correlated, the former does not cause the wind, and moist cornerstones do not directly lead to rainfall. Despite their strong association, there is no direct causal link between them. Traditional approaches to causal analysis involve methods like controlled experiments, such as randomized double-blind trials in medicine and AB testing in the IT industry. However, due to ethical, legal, or practical constraints, the controlled variable method may not be applicable in all scenarios [5].

In the realm of causal research, Turing Award laureate Professor Pearl has divided causal analysis into three tiers: the first tier investigates "association," the second tier examines "intervention," and the third tier delves into "counterfactual reasoning." Traditional machine learning algorithms, including deep learning, are positioned within the first tier, employing statistical methods for computation. While these algorithms demonstrate robust capabilities in correlation analysis, they fall short in explaining the underlying causal mechanisms driving the relationship between two variables [5]. For instance, deep learning methods can accurately model the relationship between surface soil temperature and temperature at a depth of 10cm. However, they cannot discern whether surface soil temperature drives changes in the 10cm depth temperature or vice versa [6]. Compared to correlation, causality rigorously distinguishes between "cause" variables and "effect" variables, playing an indispensable role in revealing mechanisms underlying phenomena [7].

Currently, research based on correlation, including many machine learning algorithms, is advancing rapidly, while investigations into causal relationships between variables are still in the exploratory stage [8,9]. Scholars such as Professor Pearl argued that causal research should be built upon an understanding of system processes and the construction of interpretable causal graph models, as causality cannot be precisely detected solely through data [5]. However, constructing causal graphs requires a deep understanding of the complex mechanisms behind system interactions, necessitating precise, intricate, and domain-specific knowledge representation. Some dynamic systems, typified by natural systems, exhibit openness, randomness, and complexity. They display typical non-linearity and chaotic characteristics, influenced by various factors that change over time. Consequently, constructing causal graphs is a formidable challenge [10]. Therefore, inferring causal relationships from observational time-series data in a model-free manner is a valuable research direction [11].

Empirical Dynamic Modeling (EDM) is a data-driven dynamic system modeling framework that distinguishes itself by eschewing the formalized methods typical in traditional data analysis, focusing solely on reconstructing the behavior of dynamic systems from time series data [12]. EDM is a vital research methodology for causal analysis of complex systems, with wide applications in fields such as ecology, finance, meteorology, medicine, and chemistry.

This paper primarily conducts a systematic review of causal detection research based on EDM. The subsequent sections of this paper are organized as follows: Section 2 analyzes the dialectical relationship between correlation and causality; Section 3 introduces the developmental history and core concepts of causal detection using EDM; Section 4 presents some improved methods proposed by scholars to address some different problems; Section 5 discusses practical applications based on EDM in fields like ecology, finance, and medicine; Section 6 outlines future development trends; and Section 7 concludes the entire paper.

## 2. Correlation and Causality

In everyday language, the terms dependency, association, and correlation are often used interchangeably. However, from a scientific perspective, dependency and association have similar meanings but differ in significance from correlation [3].

Association (dependency) represents a prevalent relationship, indicating that one variable provides information about another. Correlation is a more specific concept in statistics, referring to situations where two variables exhibit increasing or decreasing trends together. It is typically measured using Pearson correlation coefficients or Spearman correlation coefficients. While causality has a relationship with association (dependency), there is no inherent connection between correlation and causality.

### 2.1 Correlation ≠ Causality

The concept that "correlation does not imply causation" has had a profound impact since the 17th century [13]. As noted by the renowned American scientist Herbert

Simon, even in the first course in statistics, the slogan "Correlation is no proof of causation!" is imprinted firmly in the mind of the aspiring statistician or social scientist [14]. The regularities or correlations observed between two events or processes are neither a sufficient nor a necessary condition for the existence of a causal relationship between them. Scholars from various fields often sum up the relationship between the two as "establishing meaningful correlations is one thing, making the leap from correlation to causation is another" [15].

While there is still no universally accepted rigorous definition of causality in the academic community [16], many scholars continue to misunderstand correlation as causation. Correlation can occur without causation, leading to what is known as "spurious correlation," which is a statistical phenomenon where two variables exhibit a high correlation coefficient despite lacking a causal relationship [17]. The reasons for this phenomenon typically fall into three categories: mathematical, methodological, and causal structure. Much research focuses on "spurious correlation" resulting from causal structure, which can be explained by a simple model involving two independent samples driven by the same external factor.

For instance, based on empirical experience, there is a highly reliable correlation between barometric pressure readings and the occurrence of storms. However, this does not mean that low barometric pressure readings cause storms or that storms cause low barometric pressure readings. When a causal structure diagram is drawn, it becomes evident that there is a "latent factor" driving both occurrences simultaneously, as depicted in Figure 1. This latent factor is referred to as a confounding variable [18]. Therefore, even though there is no direct interaction between the two events, they exhibit a strong statistical correlation.

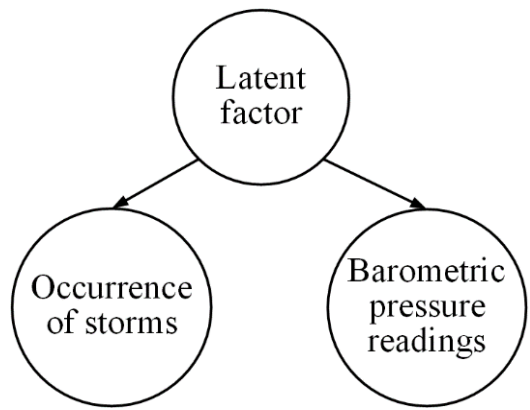

**Fig. 1** Spurious correlation due to causal structure

The Moran Effect in ecology is indeed a classic scenario where "spurious correlation" may exist [19]. Ecological systems are typically driven by external environmental variables such as precipitation and temperature. Many species share similar environments, which can lead to behavioral correlations and synchrony among non-interacting species, as depicted in Figure 2. Although there may be a strong correlation between two populations, there is no direct causal relationship between them.

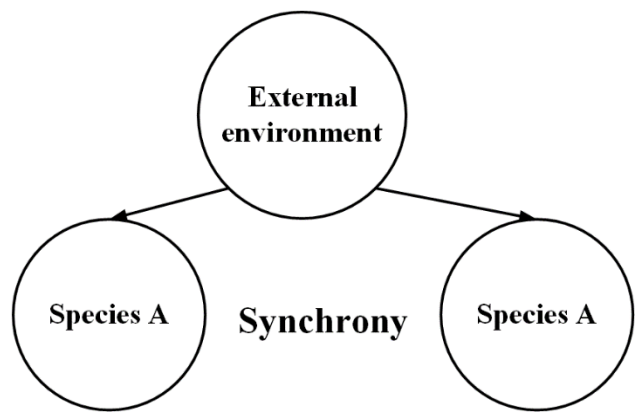

**Fig. 2** Moran Effect in ecology

This underlying causal structure indeed results in "spurious correlation" between the two variables, and traditional statistical methods often struggle to detect the presence of a third-party latent factor. Andrew Ward suggests that while knowledge of spurious correlation may have practical value, its utility depends on understanding the causal structure behind the correlation [17].

In practice, relying too heavily on conclusions drawn from the correlation between variables can often lead to mere coincidental findings. For example, a study by Radoslaw and colleagues examining residents of different floors in Swiss buildings found that residents living on the eighth floor or higher had reduced risks of cardiovascular diseases, cancer, and respiratory system diseases compared to those on lower floors [20]. However, drawing the conclusion that "living on higher floors reduces the risk of diseases" based solely on this correlation would be unreasonable. Therefore, in scientific research, relying solely on correlation for analysis has certain limitations.

## 2.2 No Correlation ≠ No Causation

Just as correlation does not imply causation, the absence of correlation does not necessarily mean there is no causal relationship. Besides the "spurious correlation" discussed earlier, in fields like ecology, when using traditional correlation statistical analyses, many true interactions may remain invisible. In dynamic systems, two variables that interact with each other nonlinearly may exhibit periodic changes in their correlation patterns, including positive correlation, negative correlation, or no correlation at different times, which can pose challenges to traditional statistical analysis [2].

For instance, in the Baltic Sea, the correlation between the populations of fish and zooplankton varies over time, sometimes showing positive coupling, sometimes negative coupling, and sometimes being completely uncorrelated. Ecologists, such as George Sugihara, refer to this phenomenon as "mirage correlations," signifying that the correlation between two interacting variables can change or disappear with shifts in the state of the dynamic system. For example, in the Lorenz butterfly attractor, two variables exhibit completely opposite correlations on different wings [21]. This state-dependent behavior is a hallmark of complex nonlinear systems and is prevalent in the natural world.

Not only in complex systems like those found in nature but even in simple nonlinear systems, "mirage correlations" can be observed. For instance, consider the following coupled difference equations:

$$X_{t+1} = 3.8X_t(1 - X_t) - 0.02X_tY_t$$

$$Y_{t+1} = 3.8Y_t(1 - Y_t) - 0.08Y_tX_t$$

Assuming initial values for X and Y as 0.2 and 0.5, respectively, analyzing the time series data of these two variables yields the results shown in Figure 3. The Pearson correlation coefficient $r$ in three-time intervals [60,70], [260,270], and [840,850] is 0.84, -0.01, and -0.93, respectively, indicating positive correlation, no correlation, and negative correlation characteristics. Despite this system being deterministic and known, it lacks a fixed correlation indicator for long-term description.

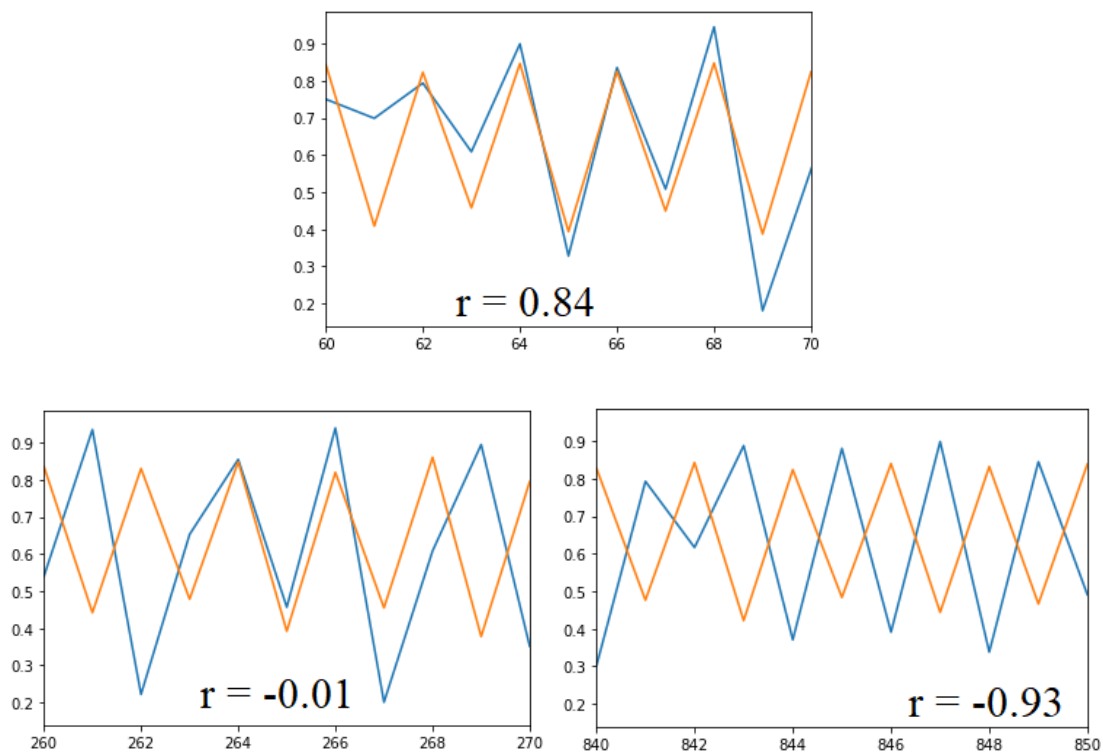

Fig. 3 Examples of Mirage Correlations

## 3. Empirical Dynamic Modeling Theory

EDM theory is a data-driven dynamic system modeling framework proposed by Professor Sugihara et al at the University of California, which was published in a paper in the Proceedings of the National Academy of Sciences (PNAS) in 2015 [12]. The most significant feature of EDM is the abandonment of the formalized methods commonly used in traditional data analysis and instead reconstructing the behavior of dynamic systems solely from time series data. Researchers applied EDM in a data-driven manner to predict the population sizes of Fraser River salmon, Pacific sardines, and Atlantic menhaden in 2014. The predictions achieved by EDM were more accurate than any previous modeling methods. In the same year, a review article by Donald and colleagues was published in PNAS, highly praising EDM theory. They believed that with the improvement of computational capabilities, data-driven modeling methods had the potential to change the dominance of traditional model equations in the study of dynamic systems and open a new research approach for analyzing complex dynamic systems [22].

EDM initially originated and was applied in ecology and later expanded to fields such as finance, chemistry, agriculture, and industry. Before EDM was introduced, many scientific fields approximated real-world situations using model equations to explain historical observations and predict future values. In most cases, these equations were based on prior knowledge, and the process of refining prior knowledge was highly complex. Additionally, the nonlinear nature of dynamic systems made traditional model equations highly sensitive to structure and parameters. As the system's state changed, model parameters also continuously changed, leading to the phenomenon of overfitting. Consequently, fully parameterizing the model was nearly impossible, and accurate predictions of the future were challenging [12]. On the other hand, nonlinearity and dynamics are prevalent in complex systems, suggesting that traditional linear methods such as principal component analysis, k-means clustering, and various regression algorithms are not suitable. These linear methods primarily rely on correlations for analysis. Therefore, for dynamic systems composed of variables with nonlinear interactions, although correlations may be implicit in heuristic thinking, they are fundamentally incorrect as an approach.

The core of EDM theory is the Takens' Embedding Theorem [23], a mathematical theory that reconstructs the strange attractor of a dynamic system from time series data. This method was originally developed by Takens and subsequently extended by other researchers [24]. EDM encompasses a series of nonlinear methods based on State Space Reconstruction (SSR), also known as delay-coordinate embedding [25]. SSR originates from the idea of non-separability in nonlinear systems, which means that the entire dynamical information of the system can be obtained from a single variable [16]. In EDM, these methods do not use any model equations but instead reconstruct the behavior of dynamic systems from time series data, inferring patterns and associations from the data without relying on model equations, making them highly flexible. These methods can describe fundamental properties of dynamic systems, including complexity, predictability, and nonlinearity, and identify causal relationships between variables in the system.

The logic of EDM is based on the following fact: dynamic systems can be described as a set of states evolving in a high-dimensional space driven by certain underlying rules over time. The process of the evolution of these states can be captured by reconstructing the states evolving over time. This high-dimensional space is often referred to as the attractor manifold $M$. The motion of the attractor manifold $M$ can be projected onto the coordinate axes corresponding to specific variables, forming time series for those variables. Therefore, any continuous set of observations of the system's state can be described as a time series. When enough time series data is collected, it becomes possible to reconstruct the system's dynamical behavior in the high-dimensional space.

For instance, let's assume we know that the dynamics of soil are influenced by soil temperature (x), soil moisture (y), and soil microbial activity (z). By plotting time series related to soil temperature, soil moisture, and soil microbial activity along the X, Y, and Z axes in a 3D state space, the original soil dynamics can be reconstructed [26]. Over time, these three variables trace the trajectory of the system's state changes, providing an intuitive view of how the system evolves over time, as shown in Figure 4. The trajectories of variables will track a 3D attractor manifold $M$ in three-dimensional phase space over time. At any moment t, $M$ reflects a set of differential equations that generate a strange attractor, and points on $M$ tend to converge towards it. The instantaneous states of the soil system can all be described using points on the attractor manifold $M$.

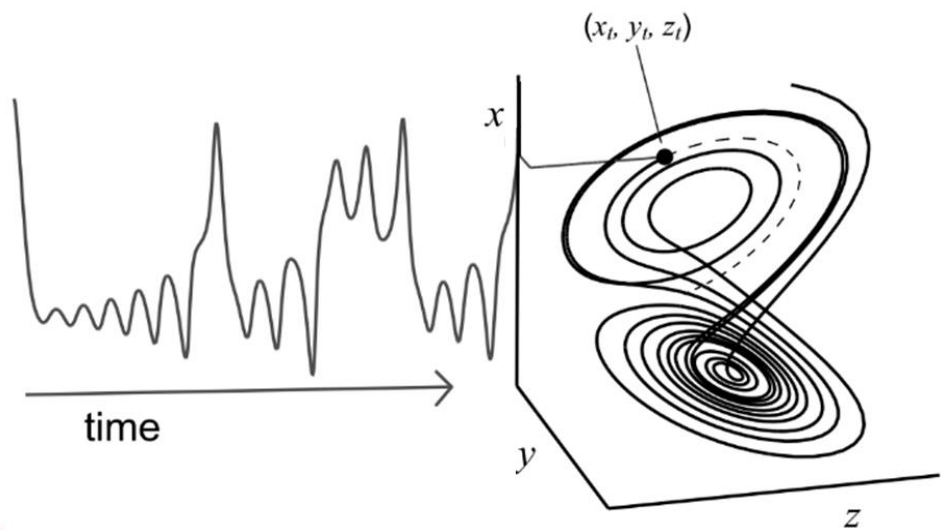

Fig. 4 State space reconstruction in soil dynamic system

In practical situations, if the fundamental equations governing the original attractor manifold $M$ are known, then it becomes relatively straightforward to characterize the complexity, nonlinearity, and causal relationships among variables in the dynamic system. For example, in the earlier example of the coupled difference equations, the complexity of the system can be described as 2, and there are bidirectional causal relationships between the two variables. However, in many cases, the differential equations governing *M* are unknown, and some variables may be challenging to quantify. For instance, the microbial activity in a soil system may be difficult to measure.

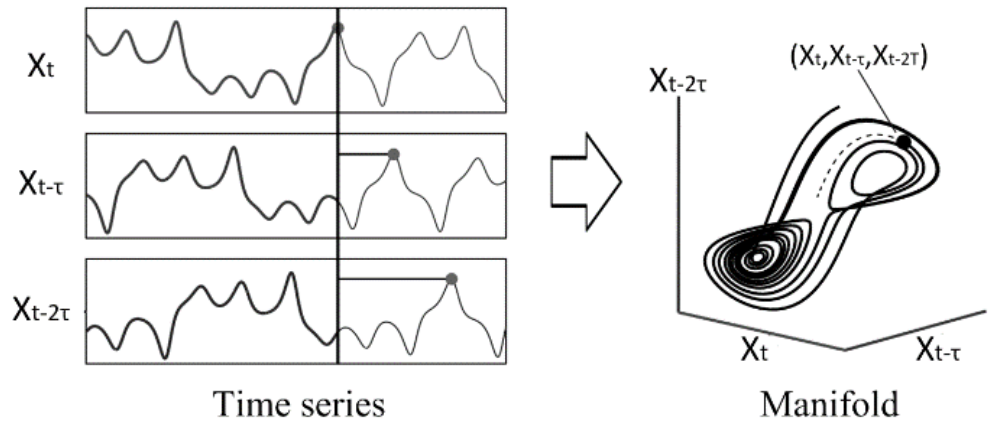

**Fig. 5** reconstruction of dynamic system through the single variable lagged coordinates

To address this issue, it is possible to use a single, sufficiently long time series of an observable variable X (such as temperature or moisture) to reconstruct the properties of the original attractor manifold *M*. This is because, if X is a projection of *M* in one direction, the Takens' theorem states that the entire deterministic development of the dynamic system can be reconstructed using the lag of a single variable X, as shown in Figure 5. Let's assume that the original time series of the variable X is $\{x_1, x_2, x_3, ...\}$. You can create a time series for X(t) by taking a segment of the time series:

$$X(t) = \{x_t, x_{t-1\tau}, x_{t-2\tau}, ..., x_{t-(E-1)\tau}\}$$

Here, the length of the sequence is E, which represents the embedding dimension. Different instances of X(t) can form an E-dimensional attractor manifold $M_x$, commonly referred to as the shadow manifold. Each X(t) represents a state point on $M_x$, and the geometric distribution properties of these points reflect the complex nonlinear interactions implicit in the time series. The optimal value of the embedding dimension E can be estimated using algorithms such as the simplex projection method. It is not necessary for the reconstructed dimension E of $M_x$ to be the same as that of $M$.

In mathematics, $M_x$ and the original $M$ are diffeomorphic, preserving properties such as Lyapunov exponents, entropy, unstable periodic orbits, and other nonlinear dynamic characteristics. The local neighborhoods of points in $M_x$ map to local neighborhoods of the original M, and based on topological invariance, $M_x$ retains the fundamental features of the original manifold M.

## 3.1 Takens' Embedding Theorem

The Takens' Embedding Theorem originated from early research on chaotic motion. Since the discovery of chaotic motion by American atmospheric scientist Lorenz in 1963, the analysis of chaotic phenomena has emerged in almost all scientific disciplines, demonstrating that many phenomena in the natural world are nonlinear. In 1980, Packard and his colleagues proposed a delay-coordinate method that uses scalar time series data to reconstruct strange attractors, which had a profound impact on subsequent research.

The Takens' Embedding Theorem states that each variable contains information about all other variables, allowing the study of a system using only a single time series. The method involves representing the time-delay coordinates of a single variable as a representation of the other variables. Simultaneously, each variable can be considered as a projection of the system's state onto a specific coordinate axis. This means that the time series of each variable represents the projection of the system's dynamics onto a specific axis, and thus, the time series of each variable contains information about the original system's dynamics.

State space reconstruction involves reconstructing a state space equivalent to the original system dynamics from a one-dimensional time series of a single variable using a delay time $\tau$ and an embedding dimension E. These parameters are crucial in the reconstruction process. If the absolute value of the delay time $\tau$ is too small, $x_t$ and $x_{t-1\tau}$ will be very close numerically, indicating strong correlation, and the embedding coordinates will essentially express the same information linearly. If the absolute value of $\tau$ is too large, it can lead to drastic changes in the system's dynamic characteristics due to the butterfly effect. Therefore, the choice of $\tau$ should ensure that $x_t$ and $x_{t-1\tau}$ are mutually independent while maintaining some statistical significance in their differences. Many researchers have conducted studies on the numerical selection of $\tau$, but in practical EDM analysis, when the system data collection is not oversampled, $\tau$ is typically set to 1. The optimal choice of embedding dimension $E$ has also been explored by many researchers. In EDM analysis, it is generally estimated using algorithms such as Simplex Projection.

## 3.2 Simplex Projection Algorithm

The Simplex Projection Algorithm, based on the reconstruction of the phase space from historical time-series data, is a method employed for short-term forecasting of chaotic dynamical systems. It is an equation-free prediction technique rooted in the concept of phase space reconstruction [33]. The underlying principle of this method is that when predicting the future behavior of a dynamic system based on past time series, optimal forecasting results are obtained by considering neighboring points [34]. This approach allows us to estimate the embedding dimension E of the attractor manifold $M$ formed by the time series data. Furthermore, the value of E in the context of EDM can characterize the complexity of the dynamical system and can also be defined as the number of variables required for reconstructing the attractor.

Due to the presence of the "butterfly effect", chaotic motion exhibits long-term unpredictability. However, in the short term, when deterministic patterns govern the system, the trajectory of the system remains confined, making chaos predictable over short horizons. If the time delay τ is relatively small, it is possible, in principle, to assume that points within the neighborhood will remain in proximity after time τ. Leveraging this neighborhood property, the Simplex Projection Algorithm estimates the prediction target through projection operators derived from the evolution of the simplex. This algorithm obviates the need to construct complex models, relying instead on historical data sequences for prediction and thereby circumventing the introduction of subjective factors inherent in modeling.

According to Takens' Embedding Theorem, the dynamical characteristics of a system can be reconstructed from the time series of a single variable, such as the time-delayed sequence $\{x_t, x_{t-1\tau}, \dots, x_{t-(E-1)\tau}\}$. During the reconstruction process, the choice of E is flexible; it need not be identical to the complexity of the original system (often referred to as the dimension of the attractor) D, nor does it need to match the number of interacting variables. In the aforementioned soil example, the assumed number of system variables is 3, while in the coupled equation example, the formula yields a variable count of 2. Nevertheless, E has an upper limit [35], and Whitney mathematically proved that E < 2D + 1. In most practical scenarios, the value of E (representing the complexity of the system) is not known a priori and requires estimation. Determining the embedding dimension E is generally the primary step in EDM analysis [36].

When employing the Simplex Projection Algorithm for time series analysis, akin to certain machine learning algorithms, data is typically partitioned into two sets: a training set, denoted as X, used for attractor reconstruction, and a prediction set, denoted as Y, employed to evaluate the prediction skill of the reconstruction model.

Here, $x_{t_1}$ represents the time series value at time $t_1$ within the training set, where an E-dimensional state point $X_{t_i} = \{x_{t_i}, x_{t_i-\tau}, \dots, x_{t_i-(E-1)\tau}\}$ can be formed from E sequential values. In the prediction phase, the same dimensionality E is applied to process the time series data within the prediction set. For a point $Y_{t_k} = \{y_{t_k}, y_{t_k-\tau}, \dots, y_{t_k-(E-1)\tau}\}$, the Euclidean distances between $Y_{t_k}$ and all points $X_{t_i}$ on the reconstructed attractor in the state space are computed. Assuming $X_{(1)}$ is the nearest neighbor through this calculation, $X_{(2)}$ is the second nearest neighbor, and so forth, up to $X_{(E+1)}$, different weights are assigned to these E+1 nearest neighbors relative to $Y_{t_k}$. These weights $W_{(i)}$ for point $X_{(i)}$ can be expressed as:

$$W_{(i)} = e^{-\frac{d(Y_{t_k}, X_{(i)})}{d(Y_{t_k}, X_{(1)})}} / \sum_{1}^{E+1} e^{-\frac{d(Y_{t_k}, X_{(i)})}{d(Y_{t_k}, X_{(1)})}}$$

In topology, these E+1 neighboring points can form a minimal polygon, known as a simplex, that encloses the prediction point $Y_{t_k}$. Therefore, in the Simplex Projection Algorithm, the number of nearest neighbors is generally chosen to be E+1. The next time point for these E+1 points, denoted as $\phi(X_{(1)}), \dots \phi(X_{(E+1)})$, is calculated as the manifold evolves over time. Predictions for $Y_{t_k+1}$ can be made, with $Y_{t_k+1} = \phi(Y_{t_k})$, and the weighted average of these E+1 points yields the prediction value $\hat{Y}_{t_k+1}$.

$$\hat{Y}_{t_k+1} = \sum_{i=1}^{E+1} W_{(i)} \phi(X_{(i)})$$

In the original algorithm, short-term predictions for future points with different time delays $T_p$ can be made, with larger $T_p$ values resulting in poorer prediction performance. In EDM analysis, $T_p$ is typically set to 1, and prediction performance is evaluated solely by comparing the correlation coefficient $\rho$, mean absolute error (MAE), mean squared error (MSE), and similar metrics between $Y_{t_k+1}$ and $\hat{Y}_{t_k+1}$. If the time series length is limited, leave-one-out cross-validation can be utilized, employing one point as the test set while using the remainder as the training set to maximize the utility of limited data. Consequently, even in cases of limited time series data, the Simplex Projection Algorithm can yield favorable results [33].

By varying the value of E, different levels of predictive performance can be computed, with the optimal embedding dimension E corresponding to the highest predictive accuracy. The Simplex Projection Algorithm has demonstrated effective predictive performance in the analysis of human disease data, ocean physics, and biological datasets [37]. It serves as a foundational algorithm within EDM theory, with some scholars extending it from single-variable embedding to multi-variable embedding and multi-view embedding, thereby enhancing its prediction skill and practicality [38,39]. Sugihara et al. further extended nearest neighbors from points to all points on the manifold, resulting in the S-map algorithm, widely applied in nonlinear system identification [40]. The concept of prediction through neighboring points on attractor manifolds has profoundly influenced the development of EDM theory [41,42].

**3.3 Convergent Cross Mapping Algorithm**

Nobel laureate Granger introduced a framework for detecting causal relationships based on predictiveness rather than correlation in 1969 [5]. This framework, generally referred to as Granger Causality (GC), assesses causal relationships in time series data by evaluating predictive capabilities. It assumes separability within the system, meaning that causes can be distinguished from effects, and that a variable can be directly removed from the system. If the predictive performance of variable X on variable Y diminishes after removing variable X, it is inferred that variable X contains information that enhances the prediction of variable Y, thereby identifying variable X as the cause of variable Y. GC is feasible in systems exhibiting separability [43] and has seen significant success in fields like economics. However, GC is not suitable for dynamic systems [44] since separability is a characteristic of purely random and linear systems, while dynamic systems may involve interactions among variables, lack separability, and often intertwine causal mechanisms within the same time series, as demonstrated in section 2.2 with coupled difference equations. In experiments, the predation intensity of peacock fish on fruit flies or worms depends on which predator type is more abundant. Thus, the predation switching behavior demonstrates a nonlinear state-dependency, making it impossible to directly remove worms from the system.

In 2012, Sugihara et al. from the University of California introduced the Convergent Cross Mapping (CCM) algorithm for causal detection in dynamic systems' time series data, published in *Science* [2]. CCM builds upon the work of numerous predecessors [45,46,47] and is based on the core idea that if variable X is the cause of variable Y (X ⇒ Y), then information from variable X must be implicitly contained in variable Y and can be recovered from variable Y.

One of the characteristics of dynamic systems is the presence of nonlinear interactions among variables, leading to chaotic phenomena, and often exhibiting non-separable, weak to moderate coupling among variables, rendering GC ineffective for causal detection. Addressing this issue, CCM assesses the strength of the causal relationship in the X ⇒ Y direction (i.e., X causing Y) based on the extent to which variable Y can be reliably inferred from variable X. CCM is a data-driven approach that does not rely on any equations or prior knowledge but infers causal relationships directly from the time series of variables. Researchers have tested CCM using real ocean data, analyzing the causal relationships among sardine abundance, northern anchovy abundance, and sea surface temperature (SST) near the coast of California, demonstrating the effectiveness of this data-driven causal detection method [2]. CCM is not contradictory to GC but serves as a complement in the context of nonlinear dynamic systems [48]. GC is suitable for purely random systems, whereas nonlinear dynamic systems are deterministic and not entirely random, with underlying rules (attractor manifolds) governing the system's evolution. In the theory of dynamical systems, if two time series variables originate from the same dynamical system, they exhibit causal associations and share an attractor manifold M, indicating that one variable can be used to identify the state of another variable.

For example, in a predator-prey model where there is a causal relationship with mutual interactions between both (predator ⇔ prey), it is possible not only to recover information about the prey from the predator's time series but also to recover information about the predator from the prey's time series. In many practical scenarios, causality is unidirectional. Assuming an environmental variable I is a random environmental driver for the population count P (I ⇒ P), it is possible to estimate I-related information from P but not vice versa. For instance, sea surface temperature (SST) can affect fish abundance, while fish abundance does not influence SST. Therefore, the time series of fish abundance can estimate the time series of SST, but the reverse is not true.

CCM detects the X ⇒ Y causal relationship by accurately estimating the historical time series of variable X based on the historical time series of variable Y, a measure that can be mathematically quantified as prediction skill $\rho$. It is typically described by the correlation coefficient between predicted values and actual values. To maintain a positive interpretation, the square of the correlation coefficient [49] or other metrics like MAE and RMSE [50] can sometimes be used. The greater the prediction skill $\rho$, the stronger the causal relationship. When a causal relationship exists, as the length of the variable's time series L increases, this prediction skill gradually increases and converges toward a stable value.

The computational process of CCM bears a striking resemblance to the Simplex Projection Algorithm. Assuming the time series of variable Y is $\{y_1, y_2, ... y\}$, and the time series of variable X is $\{x_1, x_2, ... x_L\}$. By reconstructing the attractor manifold $M_y$ for variable Y with time delays and applying the Simplex Projection Algorithm to determine the optimal embedding dimension E, the state point on $M_y$ at time t can be represented as $Y_t = \{y_t, y_{t-\tau}, ..., y_{t-(E-1)\tau}\}$. Based on the reconstructed $M_y$, a prediction can be made for $x_t$, denoted as $\hat{x}_t|M_y$. Identifying E+1 neighboring points on $M_y$ corresponding to $Y_t$ at different times $t(1), t(1)… t(E+1)$, and considering the Euclidean distances between these points and $Y_t$, weights $W_i$ are assigned using the previously mentioned weight formula. Finally, the prediction of $\hat{x}_t$ can be expressed as:

$$\hat{x}_t|M_y = \sum_{i=1}^{E+1} W_i\, x_{t(i)}$$

In cases where the time series length L is sufficiently long, and in the absence of noise, the ideal scenario results in the value of $\hat{x}_t|M_y$ converging to the true value $x_t$, at which $\rho$ tends to equal 1. This convergence phenomenon is a mathematical principle.

CCM detects causality based on prediction skill, sharing similarities with the Simplex Projection Algorithm. However, while the Simplex Projection Algorithm fundamentally predicts future values (i.e., the dynamics of the system), CCM estimates the mutually implied information between variables and does not predict the system's evolutionary rules. This distinction allows CCM to avoid information loss caused by chaotic phenomena.

### 3.3.1 Bidirectional Causality

Bidirectional causality refers to a causal relationship between two variables (X ⇔ Y) where they interact with each other, and this is a primary case discussed under the Takens Embedding Theorem. When variables are mutually coupled, they exhibit convergence phenomena in both directions within CCM.

The Takens' Embedding Theorem demonstrates that the dynamical characteristics of a system can be recovered from the time-delayed time series of a single variable. The evolution trajectories and state points of all variables contained in a system constitute the original attractor manifold M. Suppose the system contains two variables, X and Y, with a causal relationship involving mutual interaction between X and Y. The reconstructed attractor manifolds, separately based on the time delays of X and Y, are denoted as $M_x$ and $M_y$. The three manifolds, M, $M_x$, and $M_y$, are differentiable homeomorphic, preserving topological invariance [36], and there is a one-to-one mapping relationship between state points on these three manifolds.

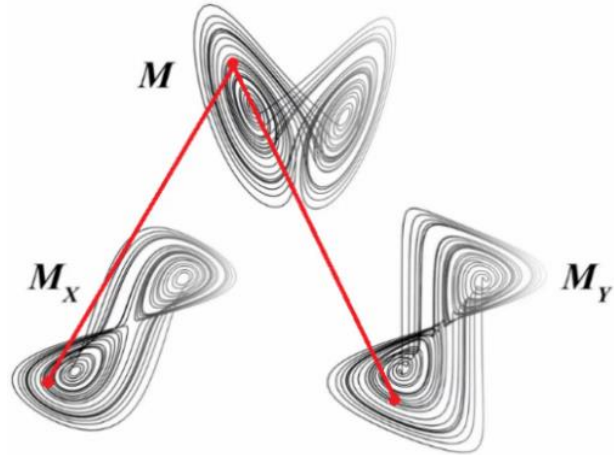

**Fig. 6** The state points have a 1-1 mapping relationship between the three manifolds

As the length of the time series used for reconstruction (L) increases, the density of the reconstructed attractor manifolds also increases, the attractor trajectories become clearer, and the accuracy of predictions based on neighboring points improves, leading to the convergence of prediction skill $\rho$ to a stable value. Convergence is a necessary condition for the presence of causality and a key attribute distinguishing causality from correlation, indicating a positive correlation between L and prediction skill $\rho$. To illustrate this convergence, one can randomly extract different lengths L from the time series to reconstruct the state space and calculate $\rho$ for various L values, with the minimum L equal to the system's embedding dimension and the maximum L equal to the length of the entire collected time series. In the case of a deterministic system without noise, as L tends to infinity, $\rho$ approaches 1. However, in practical applications, the attractor manifold $M$ constructed from time series data is only a low-dimensional approximation of the true system. Convergence will be constrained by observation errors, process noise, and the length of the collected sequences [51]. Prediction skill $\rho$ typically converges to a stable value less than 1. Moreover, a higher prediction skill $\rho$ indicates a stronger causal relationship in that direction.

Taking the example of the coupled difference equations from section 2.2, even though there exists bidirectional causality between X and Y, the causal relationship in the X ⇒ Y direction is stronger based on the parameters. When analyzed using CCM, both directions show convergence trends, but the prediction skill $\rho$ for the X ⇒ Y direction is higher, as illustrated in Figure 7.

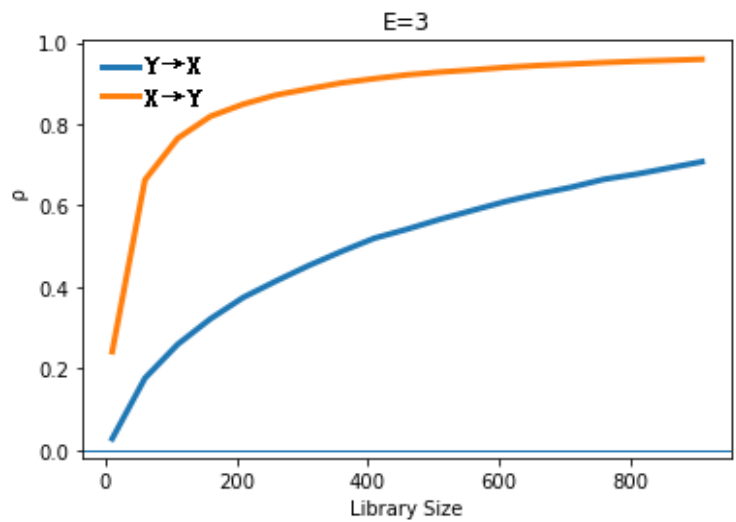

Fig. 7 CCM convergence of bidirectional coupling

### 3.3.2 Unidirectional Causality

In many practical cases, the causal relationship between two variables is not bidirectional but unidirectional, where only a single causal direction X ⇒ Y exists, meaning that variable X affects Y, while Y has no influence on X. This often occurs when X is an external environmental variable. In such cases, information flows in one direction, and it is possible to extract information about X from Y but not vice versa, which contrasts with the approach of Granger causality (GC).

In this scenario, the information flow is unidirectional. $M_y$ is an effective shadow manifold that is differentiable homeomorphic with the original $M$ and contains information about both Y and X. On the other hand, $M_x$ contains information about X only and lacks information about Y, resulting in a lack of one-to-one mapping between state points on $M_y$ and $M$ [52]. When using CCM for analysis in this direction, there is no convergence of prediction skill $\rho$, and the prediction skill $\rho$ obtained in the non-causal direction Y ⇒ X is smaller than the correct causal direction X ⇒ Y.

To illustrate this, consider modifying the coupled difference equations from section 2.2 to have a unidirectional causal relationship, removing the causal connection in the X ⇒ Y direction, and then performing the CCM analysis. The experimental results, as shown in Figure 8, reveal that there is no convergence in the X ⇒ Y direction, indicating the absence of causality in that direction. However, in the Y ⇒ X direction, there is a clear convergence phenomenon, indicating the presence of a causal relationship in that direction.

$$X_{t+1} = 3.8X_t(1 - X_t) - 0.02X_t Y_t$$
$$Y_{t+1} = 3.8Y_t(1 - Y_t) - 0.08Y_t$$

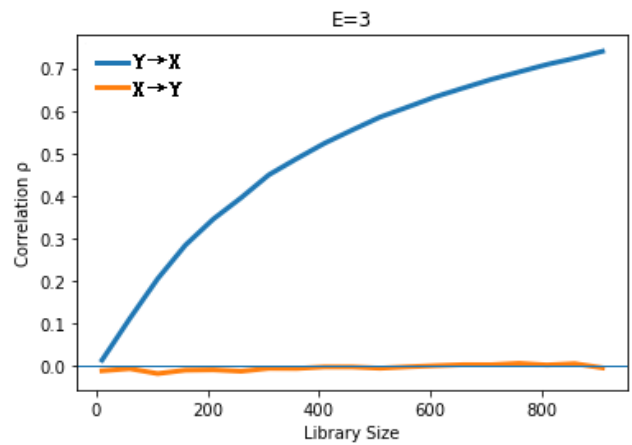

Fig. 8 CCM convergence of unidirectional coupling

A typical example of a unidirectional causal relationship is the analysis conducted by Sugihara et al. using CCM to investigate the relationship between sardine abundance, anchovy abundance, and sea surface temperature (SST) [2]. Based on the convergence of the experimental results' curves, it was found that SST has a clear causal relationship with both fish populations, sardines and anchovies. However, there is no causal relationship between the two fish populations themselves, nor do they have a causal relationship with SST. These findings align with common knowledge and intuition.

### 3.3.3 Complex Network Relationships

Due to the complexity of natural environments, systems often involve not only causal relationships between two variables but also complex network relationships. These causal networks share similarities with the Structural Causal Models (SCMs) proposed by Professor Pearl but are primarily used for analyzing time series data, whereas SCMs are mainly used to analyze large sets of statistically unordered data [18].

For instance, consider the Moran effect in an ecological system, as shown in Figure 2, where two populations, A and B, do not interact with each other but are both influenced by a common environmental variable, Z. In SCM, this can be represented as a fork structure, indicating that A and B are conditionally independent given Z. If Z is not deterministic, then A and B exhibit dependence. However, regardless of the situation, there is no cross-convergence phenomenon between A and B because there is no causal relationship between the two variables. This demonstrates that CCM can distinguish causality from correlation. On the other hand, there is convergence in the Z ⇒ A and Z ⇒ B directions, as information about the environmental variable Z can be recovered from either variable A or B.

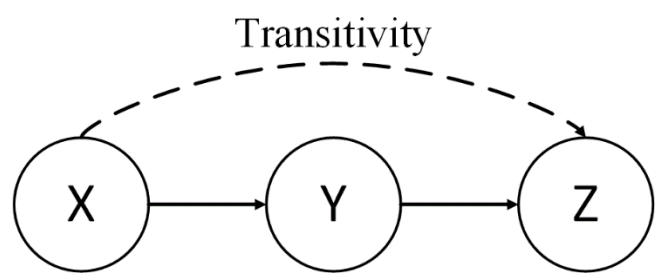

**Fig. 9** Direct causality and indirect causality

When analyzing the causal network of complex systems using CCM, causal relationships have transitivity. For example, if foxes prey on rabbits and rabbits eat grass simultaneously, there is a transitive causal relationship between foxes and grass. In other words, if X ⇒ Y and Y ⇒ Z, it implies X ⇒ Z, and if X ⇔ Y and Y ⇔ Z, it implies X ⇔ Z. Transitivity simplifies the analysis of complex network relationships using CCM.

## 4. Improvement methods

Since the inception of causal inference methods based on the EDM, numerous scholars have proposed several enhanced approaches. These enhancements predominantly focus on four key aspects: the consideration of time delays, the evaluation of predictive ability using the $\rho$ criterion, improvements in the reconstitution of manifold structures using the M method, and the judicious introduction of noise.

**4.1 Consideration of Time Delays**

Although the Convergent Cross Mapping (CCM) algorithm has been successfully applied in causal identification for various dynamic systems, Sugihara et al. observed an anomaly wherein a notably strong unidirectional causal relationship, denoted as X ⇒ Y, led to a phenomenon referred to as 'synchronization.' In this scenario, the state of the response variable Y is controlled by the state of the driving variable X, causing the entire system to 'collapse' under the influence of the driving variable X [53]. Despite Y having no causal impact on X, information about X can be obtained through Y, resulting in cross-convergence in both directions [26]. Consequently, CCM fails to accurately identify bidirectional causality and the 'synchronization' phenomenon induced by strong unidirectional driving [54].

Addressing this issue, Hao Ye et al. proposed an extended CCM algorithm [54], commonly referred to as ECCM, which stands as one of the most significant CCM enhancements. Following the general principles of causality, causal associations do not occur instantaneously; causes precede effects. By introducing different time delays and evaluating them based on the $\rho$ criterion for optimal predictive ability, ECCM can ascertain whether there exists a time delay in the driving variable's influence on the response variable. It also possesses a degree of discriminative capability for both direct and indirect causal relationships. In cases of strong unidirectional driving leading to synchronization (assuming causal relationship X ⇒ Y), this method detects negative lag in the true causal direction X ⇒ Y and positive lag in the Y ⇒ X direction. This differentiation is attributed to the response variable Y being better suited for predicting the historical values of the driving variable X, while the driving variable X is more suitable for predicting the future values of the response variable Y. ECCM thus distinguishes bidirectional causal relationships from the 'synchronization' phenomenon.

To address the computational complexity associated with ECCM, Ge et al. introduced a novel SW-ECCM algorithm. Building upon ECCM, this algorithm incorporates a sliding window technique, offering more intuitive and detailed insights into causal relationships and their strengths under different time delays. This enhancement enhances the practical value of causal identification in real-world applications [55].

**4.2 Enhancement of Predictive Assessment Criteria**

When the length of a time series, denoted as L, is relatively short, Convergent Cross Mapping (CCM) exhibits suboptimal convergence. In response to this challenge, Huanfei et al. introduced a causal detection method specifically tailored for short-term time series, known as CMS. CMS assesses causal relationships by measuring the smoothness of cross-graphs between two variables. This approach integrates neural networks to capture cross-mapping features around points on attractor manifolds, without relying on global information. Consequently, it yields effective causal detection results even with limited data volume [56].

Huanfei et al. proposed a novel CME evaluation score for CCM, designed to assess predictive capability $\rho$ by constructing a distance matrix between actual values and predicted values. This method not only enhances the accuracy and robustness of causal detection but also serves as a tool for identifying time delays between variable values in dynamic systems [57].

Daniel et al. conducted research revealing that the mapping expansion between reconstructed attractor manifolds, denoted as $M_i$, for different variables not only reflects the strength of directional coupling but also reflects the dependency of the system's time-varying states. Building upon this insight, Daniel et al. introduced a novel causal relationship index, providing a new metric for assessing the strength of causal relationships [58].

Although the Estimated Dynamic Model (EDM) is a non-equation-based modeling method, it involves parameters such as time delay $\tau$ and embedding dimension E when reconstructing attractor manifolds. Wang et al. proposed an enhanced CCM method to address parameter selection and threshold determination issues during attractor manifold reconstruction. This method is grounded in the pseudo nearest-neighbor theory to determine the optimal embedding dimension E and employs Monte Carlo simulations to evaluate convergence thresholds [59].

To address the issue of noise interference in causal prediction in certain scenarios, Emiliano et al. introduced a more robust RCCM method. This method relies on time-guided decision scores and more rigorous cross-mapping skill scores, combining CCM with information geometric causal inference methods. It is employed to resolve strong transient variable couplings and mitigate CCM's sensitivity to noise [60].

**4.3 Advancements in Manifold Reconstruction Methods**

James et al. observed instances in which CCM yielded conclusions inconsistent with intuitive notions of causality [49], often dependent on underlying system parameters. To address this, they proposed an enhancement known as PAI, introducing the concept of multivariate embedding into CCM [38]. Here, state points are defined as $X_{t_i} = \{x_{t_i}, \ldots, x_{t_{i-(E-1)\tau}}, y_{t_i}\}$, jointly estimating the value of $\hat{x}$ from the time series of variables X and Y. This innovation enhances the practicality of the CCM method.

Cummins et al. established a precise mathematical theoretical model for causal analysis within CCM and improved it using continuity testing algorithms, aligning it more closely with the underlying mathematical theory [43]. Clark et al., recognizing the challenge of working with short time series in certain ecological systems, integrated CCM with dewdrop regression and employed Bootstrapping techniques for attractor manifold reconstruction, thus enhancing causal relationship detection from limited time series data [61].

To maximize the utility of limited data for attractor manifold reconstruction, Anna et al. introduced a hybrid prediction method based on state-space reconstruction for causal inference. This approach shifts from separately reconstructing $M_x$ and $M_y$ using the time series of X and Y to jointly reconstructing the attractor manifold $M_{x+y}$ for prediction [62].

Causality possesses transitivity, and direct causal relationships reflect the intrinsic mechanisms underlying phenomena. Siyang et al. proposed a partial cross-mapping method called PCM. By reconstructing the state manifolds $M_x$, $M_y$, and $M_z$ using the time series of X, Y, and Z, respectively, and employing cross-graphs, PCM distinguishes direct causality X ⇒ Z from indirect causality X ⇒ Y ⇒ Z. Therefore,

PCM can differentiate between direct and indirect causality in complex dynamical systems [63].

Traditional EDM attractor manifold reconstruction M assumes that sample points are uniformly distributed over time. However, in practical applications, data is often irregularly sampled, which can impact the reconstruction process. Bethany et al. introduced an improved method that accommodates variable time intervals by incorporating the time intervals of sample points into the left-lagged embedding process [64].

**4.4 Rational Noise Injection**

It is commonly held that the presence of noise in data can adversely affect the accuracy of analysis. However, a counterintuitive phenomenon exists in dynamic systems. Jiang et al., based on empirical data and model studies of nonlinear ecological systems, discovered that noise can counterintuitively enhance the capacity for directional causal detection. They found that judiciously injecting asymmetric noise into time series can improve the underlying reconstruction characteristics of dynamic systems [65]. In the same year, Dan et al. also identified situations in which the CCM algorithm produced erroneous conclusions, manifesting as the inability of the predictive capability $\rho$ to exhibit a convergent trend [51]. The presence of noise in dynamic systems reduces CCM's predictive capability, and the convergence value is negatively correlated with the strength of noise. Therefore, Dan et al. argue that the rational control of noise injection can lead to more precise causal inference.

## 5. Practical Application

**5.1 Ecological Field**

The EDM theory was initially applied in the field of ecology and has seen the widest adoption in ecological research. This is because ecosystems align with the Takens embedding theorem as deterministic dynamic systems. The stability of ecological systems leads to relatively stable internal causal relationships. In the absence of external influences, interactions among species do not undergo significant changes.

In addition to the causal analysis between fish population and SST mentioned earlier, Sugihara et al. also employed CCM to detect causal relationships in predator-prey systems through the analysis of time series data. Their research revealed asymmetric bidirectional causal relationships between predators and prey [2]. Kazutaka et al. analyzed the dynamic causal interactions in insect community food chains [66]. McGowan et al. studied the dynamics of coastal phytoplankton in California, showing that irregularities in these phytoplankton populations were not random but resulted from nonlinear population dynamics driven by external stochastic factors [67]. Ye et al. investigated the nonlinear causal relationship between water level fluctuations and phytoplankton biomass in reservoirs [68]. Masayuki et al. conducted a twelve-year study on the historical data of marine fish communities in the Otsuchi Bay of Japan, constructing a dynamic interaction causal network. Their experimental results indicated weak interactions and minor population fluctuations in the summer ecosystem [69]. Cao et al. explored the causal relationships between temperature and soil moisture at different depths during the summer and winter seasons, mathematically describing the causal networks among various factors in different soil depths [6].

Chang et al. constructed causal networks for sixteen globally distributed ecosystems to gain further insights into biodiversity in aquatic ecosystems, focusing on critical factors and pathways related to species diversity [70]. Kanehiro utilized the CCM method to investigate deciduous leaf fall in tropical evergreen rainforest ecosystems, revealing that temperature was the primary driving force behind leaf fall [71]. Haiyun et al. studied the propagation of droughts, with a specific focus on the causal relationships between meteorological droughts and hydrological droughts, as well as the time lag involved in drought propagation [72]. Doi et al. used the CCM method to analyze the causal effects of seafloor particulate organic carbon and temperature on marine species richness, providing insights into deep-water circulation processes [73].

**5.2 Financial Field**

In contrast to ecological systems, financial systems are more volatile due to factors such as human disturbances [74], making it challenging to identify long-term stable causal relationships within them. However, financial time series exhibit long memory, making the use of data-driven approaches suitable for describing dynamic behavior in economic and financial systems [75].

Huffaker et al. utilized the CCM method to study the dynamic causal relationship between beer promotions and sales volume. Experimental results revealed a nonlinear coupling between the two, indicating bidirectional causality. Beer promotions significantly influenced sales volume, while sales volume also had a long-term impact on promotional decisions [76]. Azqueta conducted research on the causal relationship between news media narratives and cryptocurrency prices. The experimental findings indicated a strong bidirectional causal relationship between the two [77]. Ge et al. applied their proposed SW-ECCM algorithm to the financial system, investigating the causal relationships among returns on indices such as the Dow Jones Industrial Average and the Shenzhen Composite Index in both the Chinese and American stock markets. Researchers found that causality in stock markets is time-varying and highly dependent on unpredictable political events. These studies contribute to a deeper understanding of time-varying causal relationships in the Chinese and American stock markets [55].

**5.3 Meteorological Field**

Meteorological systems share similarities with ecological systems, but they exhibit broader influences, often subjected to global or even extraterrestrial factors. Consequently, they exhibit higher dimensions and prominent chaotic characteristics [78].

Tsonis et al. analyzed the causal relationship between galactic cosmic rays and annual variations in global temperatures. The experimental results suggested that, in the long term, there is no strong causal association between galactic cosmic rays and global temperature changes, resolving a long-standing controversy in the academic community [79]. Wang et al. employed CCM to study the sensitivity of the carbon cycle to tropical temperature changes [80], analyzing the causal relationship between the growth rate of atmospheric carbon dioxide and annual average surface temperatures. Egbert et al. studied the nonlinear features of climate dynamics, analyzing the causal relationship between temperature changes and greenhouse gas concentrations [81]. Wang et al. analyzed the causal relationship between soil moisture in the mid-low latitudes of the Northern Hemisphere and precipitation, elucidating the strength and time lag of soil moisture's impact on precipitation [82]. Huang et al. addressed the high-dimensional and chaotic nature of atmospheric systems by proposing a time-delayed CCM to detect causality in the complex coupling between the troposphere and stratosphere, analyzing causal chains triggered by polar vortex activity [83]. Zhang et al. used CCM to analyze the causal relationships and feedback effects between the Siberian High, winter surface temperatures, and the North Hemisphere circulation pattern in the Northeast Asia region, offering insights into the analysis of seasonal climate variations [84].

**5.4 Medical Field**

The EDM theory also holds significant application value in the medical field, where complex physiological processes within the human body and the spread of various diseases can be viewed as dynamic systems. The nonlinear, dynamic, and chaotic long time-series data provide rich scenarios for causal detection using EDM [79].

Heskamp et al. applied the CCM method to assess cerebral autoregulation based on time-series data of arterial blood pressure and cerebral blood flow velocity, which has significant practical value for bedside monitoring [85]. Joseph et al. employed the CCM method to develop novel biomarkers to distinguish between normal aging, mild cognitive impairment, and early Alzheimer's disease, capturing biological activity characteristic changes caused by cognitive deficits [86]. Ethan et al. used the CCM method to study global environmental drivers and critical thresholds for influenza outbreaks through epidemiological time series data, revealing that absolute humidity and temperature are contributing factors to global influenza outbreaks [87]. Martin et al. investigated human cardiovascular time series data using nonlinear methods, providing important insights into complex physiological interactions such as cardiorespiratory coupling and neuro-cardiac coupling [88]. Ajay et al. studied the causal mechanisms within the human cardiovascular-postural-musculoskeletal system, holding potential clinical utility in detecting falls among the elderly [5]. The transmission of infectious diseases involves complex dynamic processes that are challenging to test with fitted models. Cobey et al. analyzed historical data to examine the causal relationships between infectious diseases such as childhood polio, mumps, chickenpox, and whooping cough [90]. Ma et al. studied hospital admission records for cardiovascular diseases in Hong Kong and long-term historical air pollution data, revealing the major air pollutants responsible for the diseases and uncovering hidden time delays [57].

### 5.5 Other Applications

Dynamic systems with time-varying internal couplings are prevalent in the natural world. While the complete control equations for these systems are often unknown, there is typically a dominant element [91]. Amir et al. applied CCM to identify the dominant elements in complex coupled dynamic systems, providing valuable insights for the application of CCM in various dynamic fields.

In chemical processes, identifying the causes of faults and disturbances is crucial for enhancing safety. The complex, dynamic, and nonlinear interactions among various substances in chemical processes provide scenarios for EDM analysis. Luo et al. employed an improved CCM method to determine the paths and time delays of disturbance propagation in the Tennessee-Eastman chemical process [92]. Wang et al. studied the causal relationships among various temperature parameters during the reaction process in hydrocracking processes [59]. Xiang et al. proposed a causal analysis multi-view embedding method to predict trends in key parameters of chemical processes using small datasets, thereby enhancing production safety [93].

## 6. Research Outlook

Causal detection based on empirical dynamic modeling (EDM) has become an important research direction in the field of causal analysis. Further research in this direction primarily includes the following aspects:

(1) While the application of causal analysis theory based on EDM has been widely adopted in various research fields, it fundamentally remains a method based on pure mathematical principles. It lacks physical explanations for specific processes. Future research should focus on improving the interpretability of this method, particularly the trends in predictive capability $\rho$. Integrating this method with structural causal models could be a vital direction.

(2) Various studies have indicated that this method is more applicable in systems with fewer variables. Over long-term historical processes, dynamic systems continually evolve, making modeling less stable when there are more variables. Traditional state space reconstruction theories struggle to capture the underlying patterns in time series data, such as the changes in fish populations over time, where the same region in 2015 may differ significantly from that in 1950. Combining this method with time series analysis in machine learning to explore the inherent patterns in evolving time series data presents substantial research potential.

(3) Due to the fundamental reliance on state space reconstruction theory, this method requires long time series data for predictive capability to converge. However, obtaining extended data is challenging due to various factors such as data collection, experimentation, and technology. Improving effectiveness in small-sample data through data generation and enhanced sampling methods holds theoretical and practical significance.

(4) When data sparsely covers the state space, resulting in significant data gaps or irregular sampling, this method's causal detection performance deteriorates. State space reconstruction theory assumes that sample points are evenly distributed over time, with a fixed sampling delay time τ. Irregular data collection and sampling, a common occurrence in fields like ecology, medicine, and atmospheric sciences, pose a challenge. Research is needed to enhance causal detection capabilities in the presence of data gaps and improve the effectiveness of state space reconstruction.

(5) While some methods can distinguish between indirect and direct causal relationships in complex systems, they demand substantial data and are constrained by multiple factors, resulting in suboptimal practical outcomes. So far, no systematic and comprehensive method has been developed to address this issue effectively, leaving significant room for future research in this field.

## 7. Conclusion

Understanding the interactions among variables in complex systems is essential for comprehending the mechanisms governing system behavior. Causal detection based on empirical dynamic modeling (EDM) is a crucial method for the analysis of complex systems. This article first examined the dialectical relationship between correlation and causation, comprehensively introduced the core concepts of causal detection based on EDM, discussed some improvements in EDM theory, practical causal detection applications, and identified some current issues. Finally, it presented future development trends in this field.